\documentclass[10pt, conference]{IEEEtran}
\IEEEoverridecommandlockouts
\usepackage{cite}
\usepackage{amsmath,amssymb,amsfonts}
\usepackage{graphicx}
\usepackage{textcomp}

\newcommand*{\rom}[1]{\expandafter\@slowromancap\romannumeral #1@}
\makeatother
\usepackage{algorithm, algorithmic}
\usepackage{mathtools}
\usepackage{amsmath}
\usepackage{amssymb}
\usepackage{cite}
\usepackage{threeparttable}
\usepackage{calc}
\usepackage[table]{xcolor}

\usepackage{textgreek}
\usepackage{booktabs}
\usepackage{siunitx}
\usepackage{array}
\usepackage{balance}

\newcolumntype{L}{>{\phantom{$\mathbin{-}$}$}l<{$}}

\newcolumntype{M}[1]{>{\centering\arraybackslash}m{#1}}
\newcolumntype{N}{@{}m{0pt}@{}}
\usepackage{url}
\usepackage{hyperref}
\usepackage{makecell}
\usepackage{enumitem}
\usepackage{ragged2e} 
\usepackage{multirow}

\def\BibTeX{{\rm B\kern-.05em{\sc i\kern-.025em b}\kern-.08em
    T\kern-.1667em\lower.7ex\hbox{E}\kern-.125emX}}
\begin{document}

\title{MuaLLM: A \underline{Mu}ltimod\underline{a}l \underline{L}arge \underline{L}anguage \underline{M}odel Agent for Circuit Design Assistance with Hybrid Contextual Retrieval-Augmented Generation\\

\author{Pravallika~Abbineni\IEEEauthorrefmark{1}, Saoud~Aldowaish\IEEEauthorrefmark{1}, Colin~Liechty\IEEEauthorrefmark{1}, Soroosh~Noorzad\IEEEauthorrefmark{1},
Ali Ghazizadeh\IEEEauthorrefmark{2}, Morteza~Fayazi\IEEEauthorrefmark{1}
\thanks{\IEEEauthorrefmark{1}Department
of Electrical and Computer Engineering, University of Utah, Salt Lake City, UT, 84112 USA. \IEEEauthorrefmark{2}Department of Electrical Engineering and Computer Science, University of Michigan, Ann Arbor, MI, 48109 USA. (e-mail: u1475870@umail.utah.edu, u1275778@utah.edu, u1485690@utah.edu, soroosh.noorzad@utah.edu, alighazi@umich.edu, m.fayazi@utah.edu).}}
}

\maketitle
\begin{abstract}
Conducting a comprehensive literature review is crucial for advancing circuit design methodologies. However, the rapid influx of state-of-the-art research, inconsistent data representation, and the complexity of optimizing circuit design objectives (\textit{e.g.} power consumption) make this task significantly challenging. Traditional manual search methods are inefficient, time-consuming, and lack the reasoning capabilities required for synthesizing complex circuits. In this paper, we propose MuaLLM, an open-source multimodal Large Language Model (LLM) agent for circuit design assistance that integrates a hybrid Retrieval-Augmented Generation (RAG) framework with an adaptive vector database of circuit design research papers. Unlike conventional LLMs, the MuaLLM agent employs a Reason + Act (ReAct) workflow for iterative reasoning, goal-setting, and multi-step information retrieval. It functions as a question-answering design assistant, capable of interpreting complex queries and providing reasoned responses grounded in circuit literature. Its multimodal capabilities enable processing of both textual and visual data, facilitating more efficient and comprehensive analysis. The system dynamically adapts using intelligent search tools, automated document retrieval from the internet, and real-time database updates. Unlike conventional approaches constrained by model context limits, MuaLLM decouples retrieval from inference, enabling scalable reasoning over arbitrarily large corpora. At the maximum context length supported by standard LLMs, MuaLLM remains up to 10x less costly and 1.6x faster while maintaining the same accuracy. This allows rapid, no-human-in-the-loop database generation, overcoming the bottleneck of simulation-based dataset creation for circuits. To evaluate MuaLLM, we introduce two custom datasets: RAG-250, targeting retrieval and citation performance, and Reasoning-100 (Reas-100), focused on multistep reasoning in circuit design. MuaLLM achieves 90.1\% recall on RAG-250, highlighting strong multimodal retrieval and citation accuracy. On Reas-100, it reaches 86.8\% accuracy, demonstrating robust reasoning capabilities on complex design queries.
\end{abstract}

\begin{IEEEkeywords}
Circuit design automation, LLM, multimodal design assistant, RAG, agentic workflow, Reasoning and Act (ReAct) framework, database updating, scalable, open-source.
\end{IEEEkeywords}

\section{Introduction}
The rapid evolution of circuit design introduces numerous challenges for researchers and engineers, particularly when it comes to effectively retrieving, and utilizing vast volumes of technical information spread across various research papers. Developing innovative circuit design methodologies requires an extensive literature review. However, the traditional manual search process is both inefficient and time-consuming. It is further complicated by inconsistent data representation formats and the complexity of optimizing design objectives such as area and power~\cite{fayazi2022fascinet}. These challenges slow down circuit design innovations.

Large Language Models (LLMs)~\cite{radford2019language} have emerged as powerful tools for addressing the aforementioned challenges. As intelligent and fast conversational agents, LLMs provide context-aware insights and real-time assistance, enhancing productivity and decision-making. However, deploying LLMs in technical domains such as circuit design presents unique issues. Conventional LLMs often generate responses that lack domain-specific relevance. This matter gets exacerbated because of the hallucinating~\cite{huang2024survey} nature of LLMs generative process. This leads to outputting incorrect information, particularly when dealing with technical abbreviations or nuanced design requirements. A common mitigation strategy is to provide the model with the entire set of relevant documents as context during inference, which can help reduce hallucinations. However, this brute-force method leads to significant drawbacks. 1) It is constrained by fixed context window sizes~\cite{anthropic2023promptlength, openai2024gpt4o}. 2) It incurs high computational and memory costs, and becomes increasingly impractical as the number of papers grows, which limits scalability and efficiency. Moreover, while LLMs excel at generating coherent text, they often struggle with reasoning through iterative tasks or complex processes~\cite{gendron2023large}. This shortfall necessitates a systematic approach to harness the potential of LLMs effectively in circuit design workflows.

To overcome these challenges, this paper presents MuaLLM, an open-source\footnote{\url{https://anonymous.4open.science/r/MuaLLM-1472/README.md}} intelligent design assistant capable of answering complex, multi-step circuit design questions using a reasoning-driven approach. The proposed system addresses the aforementioned limitations through the following:

\begin{enumerate}[left=2pt]
    \item \textbf{Reasoning and Act (ReAct)-based framework:} Designing circuits involves multiple steps requiring logical reasoning, iterative actions, and decision-making~\cite{fayazi2020applications}. As discussed, traditional LLMs, which operate in a passive question-answer mode, are inadequate for these requirements. The ReAct-based framework~\cite{yao2022react} introduces a more dynamic and interactive workflow, allowing the system to reason about tasks and execute them through an iterative loop of action and feedback. This capability is critical for automating circuit design optimization and managing workflows with complex requirements.
    \item \textbf{Context-aware hybrid RAG:} To address domain-specific queries, traditional RAG~\cite{lewis2020retrieval} frameworks enhance LLMs by retrieving relevant documents from both structured and unstructured data sources. These frameworks rely on either sparse retrieval (keyword-based matching) or dense retrieval (semantic matching) techniques to identify relevant content~\cite{mandikal2024sparse}. This content is then passed as context to the LLM for generating answers to the query which mitigates hallucinations. However, these approaches often fall short in highly technical fields, including circuit design, due to their limitations in handling the complexity and specificity of such domains because: a) Sparse retrieval systems~\cite{formal2021splade} lack semantic depth and often fail to align with the nuanced terminology in technical queries; b) Dense retrieval systems~\cite{robertson2009probabilistic} may overlook rare but critical domain-specific terms that are pivotal for precise technical solutions; c) Modal limitations prevent effective integration of both textual and visual data, such as technical graphs, tables, and circuit schematics, which are crucial for solving circuit design challenges. To address these limitations, the proposed hybrid RAG framework combines the strengths of both sparse and dense retrieval techniques. Sparse retrieval excels in capturing domain-specific terminology, while dense retrieval ensures semantic relevance. This hybrid approach is further augmented by multimodal capabilities, enabling the system to effectively retrieve and generate responses based on both textual and visual data. By leveraging continuous updates, MuaLLM provides precise, contextually relevant, and up-to-date information tailored to the circuit design workflow.
\end{enumerate}

The main contributions of this paper are summarized as follows:
\begin{itemize}[left=2pt]
\item Agentic workflow: Developing a ReAct-based workflow with reasoning capabilities to automate and optimize the circuit design procedures, leveraging its ability to decompose complex problems into manageable steps. This approach provides circuit designers with real-time support, and improved decision-making, ensuring efficiency in addressing intricate design challenges.
\item Multimodal search capabilities: Enabling efficient retrieval of both textual and visual information (circuit schematics, tables, and graphs), enhancing the accessibility and usability of technical data.
\item Enhanced context-aware hybrid RAG: Designing a hybrid system combining sparse and dense retrieval techniques to provide precise, context-aware responses for text and image-based circuit design-related queries. Since MuaLLM retrieves only the most relevant chunks via a hybrid RAG pipeline, decoupling inference from corpus size making MuaLLM a scalable approach. At the maximum context length supported by standard LLMs, MuaLLm remains up to \textbf{10x} less costly and \textbf{1.6x} faster while maintaining the same accuracy. 
\item Customized tools: Incorporating custom-implemented tools into the ReAct agentic workflow such as an intelligent search engine, dynamic database updater, and netlist generator to support real-time decision-making. These tools enable rapid, no-human-in-the-loop database generation, which addresses one of the main bottlenecks in circuit design: the time-consuming process of providing labeled data through simulations.
\item Evaluation: Validating the performance of MuaLLM using two custom datasets: a) RAG-250 for assessing retrieval and citation capabilities; b) Reas-100 for evaluating multi-step reasoning in circuit design. The results show that MuaLLM achieves \textbf{90.1\%} recall on RAG-250 and \textbf{86.8\%} accuracy on Reas-100. Both datasets, along with codes, are open-sourced to encourage reproducibility and future research.
\end{itemize}

\section{Background}

The manual design of circuits is highly time-consuming and prone to human error, creating a need to accelerate the design process for faster time-to-market Integrated Circuits (ICs)~\cite{ajayi2020fully}. Artificial Intelligence (AI) algorithms have demonstrated promising results in automating circuit design, as they can learn patterns without direct human intervention~\cite{fayazi2020applications}. Among the prominent AI-based techniques used for this purpose are Bayesian Optimization (BO), Deep Neural Networks (DNNs), Graph Neural Networks (GNNs), and Reinforcement Learning (RL)~\cite{lyu2017efficient, fayazi2023funtom, dong2023cktgnn, settaluri2020autockt}.

Despite the effectiveness of these methods, they either have a slow runtime or require a large amount of labeled training data, limiting their adaptability in high-dimensional, rapidly evolving design spaces. AnGeL~\cite{fayazi2023angel} has significantly reduced both the runtime and the amount of required training data; however, it still demands extensive manual effort to provide its static labeled training set. Our work shifts focus to an earlier but critical part of the design process: enabling faster iterations through literature-driven question answering and reasoning with LLMs.

LLMs offer new opportunities for circuit design automation by introducing dynamic knowledge retrieval and reasoning. Masala-CHAI~\cite{bhandari2024masalaCHAI}, Auto-SPICE~\cite{bhandari2024autospice}, and AMSNet~\cite{tao2024amsnet} automate circuit netlist generation from schematics, reducing reliance on static datasets and enhancing design efficiency. Moreover, there are multiple works that have focused on circuit design generation using LLMs. ADO-LLM~\cite{yin2024adoLLM} improves BO by infusing domain expertise, while \textsc{CircuitSynth}~\cite{vijayaraghavan2024circuitsynth} leverages LLMs for automated circuit topology synthesis. The GLayout framework~\cite{hammoud2024human} enables LLM-driven analog layout generation using RAG, while Artisan~\cite{chen2024artisan} employs tree-of-thoughts and chain-of-thoughts approaches to improve operational amplifier design~\cite{wei2022chain, yao2023tree}. Also, LaMAGIC~\cite{chang2024lamagic}, AnalogXpert~\cite{zhang2024analogxpert}, Atelier~\cite{shen2025atelier}, and LADAC~\cite{liu2024ladac}, further reflect the rapid expansion of LLM-based approaches in circuit design.

Hybrid approaches that integrates RAG with LLMs improve knowledge retrieval and reduce hallucinations. Ask-EDA~\cite{shi2024askEDA} combines RAG with abbreviation de-hallucination, improving accuracy and recall in circuit design queries. While effective for downstream design tasks, these approaches do not fully support generalizable, literature based exploration with multimodal reasoning.

General-purpose deep research frameworks~\cite{jin2025searchr1, jiang2025deepretrieval} exist but are not well-suited for literature survey and question-answering based tasks in the circuit design domain~\cite{kamp2023open}. Furthermore, many LLM approaches lack robust multimodal reasoning and domain-specific optimization needed for handling complex design queries with multiple constraints. MuaLLM bridges these gaps by integrating a hybrid RAG framework with a multimodal LLM agent that processes the data in all the modalities (text, images, tables, and graphs). This enables efficient knowledge retrieval, iterative reasoning through a ReAct workflow, and adaptive decision-making, significantly improving circuit design automation, and scalability.

\section{Methodology}
\begin{figure*}[t]
    \centering
    \includegraphics[width=1\textwidth]{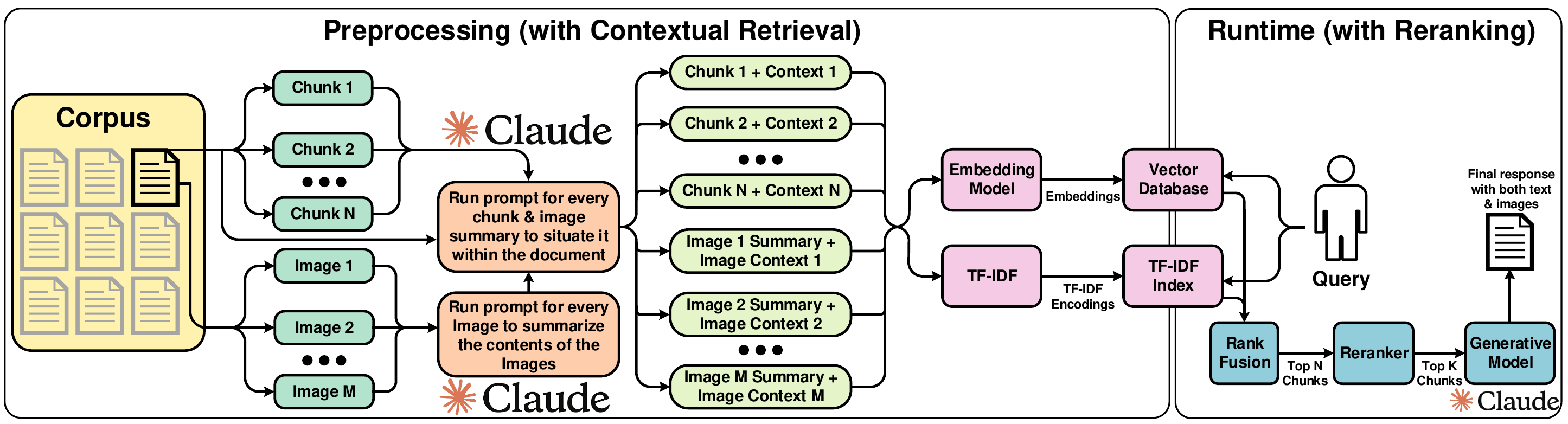}
    \caption{The proposed contextualized multimodal RAG workflow.}
    \label{fig:framework}
\end{figure*}

\subsection{Database}
The database serves as the foundation for RAG systems. Research papers in the field of circuit design are challenging to integrate into databases due to their unstructured and multimodal nature. To overcome these challenges, MuaLLM develops a robust preprocessing pipeline that structures and embeds text and images into a unified vector database, enabling precise and contextually enriched retrieval tailored to the domain as shown in Fig.~\ref{fig:framework}.

\subsubsection{Preprocessing text and images}
To preprocess the research papers, MuaLLM extracts textual data and images from PDF documents using the \texttt{unstructured} library~\cite{unstructured2025}. The extracted text is cleaned to remove artifacts and segmented into overlapping chunks. MuaLLM uses paragraph-level chunking as it gives us better search results in comparison with character-level chunking~\cite{pinecone_chunking_2025}. Overlapping chunking ensures the semantic continuity is preserved across related sections, maintaining the flow of information. This step is crucial, as processing entire documents is computationally expensive and constrained by the token limits of LLMs. Simultaneously, images are extracted and stored for further processing. These images, ranging from circuit diagrams to plots, are critical for multimodal analysis.

\subsubsection{Embeddings}
Embeddings enable efficient similarity-based retrieval by mapping data into a vector space. While initially  OpenAI’s CLIP (Contrastive Language-Image Pretraining) model~\cite{radford2021learning} was used for multimodal embedding, its general-purpose training limited its ability to handle the domain-specific details found in circuit design. 

To overcome this limitation, MuaLLM adopts a descriptive embedding approach for images~\cite{riedler2024beyond}. Each image is processed using LLM to generate detailed descriptions, capturing: a) The type of visual content (\textit{e.g.} ``circuit diagram", ``plot", or ``specification table"); b) Key elements, such as labeled electronic component types (\textit{e.g.} resistor or capacitor); c) Domain-specific technical details (\textit{e.g.} the circuit functionality of the circuit schematic image).

The enriched descriptions are embedded using the \texttt{voyage-2}~\cite{voyage22024} model and stored alongside textual embeddings in the vector database. Each embedding is paired with metadata, including the path to the original image. This setup ensures that when an embedding is retrieved, the corresponding image provides the necessary context for the generative model.

While it is true that large language models can hallucinate, our use of LLM-generated text interpretations is not for factual reasoning, but strictly to enhance search recall and indexing of visual content. These interpretations are embedded alongside original image captions using the Voyage-2 model to improve multimodal retrieval accuracy. Importantly, we evaluate the downstream impact of these embeddings on search performance using our custom benchmark datasets (RAG-250 and Reas-100). Thus, even if occasional hallucinations occur in the LLM-generated image descriptions used for embedding, they do not impact the agent's reasoning or final outputs, as the embeddings are solely intended to enhance image-aware search relevance.

\subsubsection{Contextualization}
Traditional RAG systems~\cite{lewis2020retrieval} divide documents into smaller chunks for efficient retrieval. While computationally effective, this approach often isolates chunks from their broader context, leading to ambiguous results. This lack of context can diminish the utility and reliability of the retrieved information~\cite{anthropic2024contextualretrieval}.

To generate context and improve response relevance, MuaLLM implements a contextual caching mechanism~\cite{anthropic2024contextualretrieval}. This caching system stores reusable context data generated during previous queries, minimizing redundant Application Programming Interface (API) calls and reducing computational costs.

\subsubsection{Unified vector database}
All preprocessed text and images, along with their enriched embeddings and metadata, are stored in a unified vector database. This structure facilitates seamless multimodal search, allowing effective querying of both text and images. By preserving context, the system ensures that isolated chunks or images are retrieved with their full meaning which significantly enhances relevance and usability for technical queries.

\subsection{Contextual hybrid RAG}
RAG~\cite{lewis2020retrieval} improves model responses by integrating external knowledge through a combination of data retrieval and response generation. In this process, the model first retrieves relevant information from an external knowledge base using either semantic search or keyword search. This ensures that the most pertinent and useful data is selected based on the query. After retrieval, this information is passed to a generative model as context, which then processes it to generate accurate and contextually relevant responses. The detailed implementation of the hybrid RAG is as follows.
\subsubsection{Dual search}

The hybrid contextual RAG~\cite{anthropic2024contextualretrieval} system utilizes two distinct retrieval mechanisms: semantic search and keyword search. 

\begin{itemize}[left=2pt]
\item Semantic search: This method is designed to retrieve documents based on the meaning or underlying concept of the query. It excels in situations where the user's query may involve synonyms or abstract terms. This allows the model to fetch documents that are contextually related but may not have an exact keyword match. The embeddings facilitate this semantic search, ensuring that the retrieved documents are conceptually aligned with the user’s intent.
\item Keyword search: In contrast to semantic search, BM25~\cite{robertson2009probabilistic}, a keyword-based search algorithm, focuses on exact matches of query terms within the document corpus.
\end{itemize}

By utilizing both search methods in parallel, the hybrid RAG system effectively captures both conceptually relevant and keyword-specific results, thereby expanding the breadth of retrieved data. This approach is particularly advantageous for circuit design, which is rich in both technical and semantic terminology.

\subsubsection{Ranking}
Once the semantic and BM25-based results are retrieved, the system combines the outputs from both methods to form a unified set of relevant documents~\cite{jindal2014review}. To prioritize the most relevant documents, each retrieval result is assigned a weighted score based on the search method used to retrieve it. The semantic weight typically holds more significance when context and conceptual relevance are more important, whereas the BM25 weight is employed for keyword-matching results.

\subsubsection{Re-ranking}
The re-ranking step further refines the retrieval results~\cite{anthropic2024contextualretrieval}. After the initial retrieval phase, the results are re-ranked using the \texttt{Cohere}~\cite{cohere_rerank_2024} model to reorder the documents based on their true relevance to the query. The re-ranking process leverages \texttt{Cohere}'s ~\cite{cohere_rerank_2024} understanding of the relationship between the query and the documents, ensuring that the final output is textually and semantically aligned.

\subsubsection{Response generation}
Once the relevant documents have been retrieved and refined, they are passed to the generative model (\textit{e.g.} \texttt{GPT-4o}~\cite{openai2024gpt4o}), which synthesizes the information to generate a response. The generative model uses the external knowledge extracted during the retrieval process to fill in any gaps within its internal knowledge. This creates a response that is both accurate and contextually appropriate.

While \texttt{GPT-4o} is used in our primary implementation, the system is modular and model-agnostic: \texttt{Claude 3.5 Sonnet}~\cite{claude3haiku2024} or other frontier LLMs can be easily swapped in for generation. This flexibility allows MuaLLM to adapt to evolving model capabilities, cost structures, or domain-specific strengths depending on deployment requirements.

\subsubsection{Multimodality}
An important extension of the hybrid RAG system is its ability to handle multimodal content.  This means that it can process not only textual documents but also images, graphs, and tables. By distinguishing between text and image content during the retrieval process, the system ensures that all types of data are handled appropriately. This feature is especially important in fields such as circuit design, where visual content plays a significant role alongside text. The hybrid contextual RAG system integrates the best features of the retrieval-based and generation-based approaches. By combining semantic, keyword-based search, and contextual understanding, it generates highly relevant and accurate responses.

\subsection{Agentic Workflow}
The ReAct-based agentic workflow~\cite{yao2022react} enables iterative problem-solving by cycling through \textit{thought}, \textit{action}, \textit{pause}, and \textit{observation} steps. Unlike standard LLMs that passively generate answers, the agent actively reasons through the query, determines the necessary information, and selects the appropriate tools. During the \textit{action} step, the model retrieves data using the hybrid RAG system, queries databases or employs multimodal tools as required. After pausing for results, it observes the retrieved data, reassesses gaps, and, if necessary, selects additional tools, such as fetching external documents or updating the database, to refine its understanding. This cycle repeats until the agent has gathered all the necessary information to answer the query. This ensures a more context-aware, accurate, and adaptive response compared to static LLM-based retrieval.

\subsection{Tools}
These custom-implemented tools are integrated into the ReAct-based agentic framework of MuaLLM to enable intelligent, multimodal task execution in real-time. Each tool is invoked dynamically during the reasoning process, allowing the agent to adaptively retrieve, enrich, and utilize information for circuit design queries.

\subsubsection{Database Searcher (\textit{search\_db})}
This tool retrieves relevant information from large-scale structured and unstructured circuit design literature. It supports both sparse keyword-based retrieval using BM25 and dense semantic search using vector embeddings. Results are re-ranked using the Cohere reranker, enabling high-precision multimodal retrieval of both textual content (e.g., parameter tables) and visual content (e.g., circuit schematics). The tool operates with low latency and is optimized for recall and ranking quality across hybrid search workflows.

\subsubsection{Automatic Paper Fetcher (\textit{paper\_fetcher})}
The \textit{paper\_fetcher} tool enables the agent to autonomously fetch academic literature from public repositories such as Google Scholar and arXiv. When queried about papers not present in the local database, the agent invokes this tool to download the PDF files, which are then passed to the preprocessing pipeline. This allows MuaLLM to dynamically expand its knowledge base and access the latest research with no human intervention.

\subsubsection{Dynamic Database Updater (\textit{search\_db --load\_data})}
After new papers are fetched, the \textit{search\_db --load\_data} tool preprocesses the documents and updates the vector database in real-time. It extracts text and images from PDFs, generates enriched descriptions for visual content using LLM-based captioning, and embeds the content using the Voyage-2 model. This tool ensures the database remains up-to-date, facilitating continuous learning and accurate response generation for future queries.

\subsubsection{Netlist Generator}
This tool automates the conversion of schematic circuit diagrams into SPICE-compatible netlists. It employs a YOLO-based object detection model to identify circuit components and OpenCV’s connected components analysis to extract nodes. The detected elements are parsed to form structured netlists, enabling simulation-ready outputs and eliminating the need for manual circuit annotation. The netlist generator is especially valuable for accelerating dataset creation and validating extracted designs.

\subsubsection{Netlist Generator} Converts circuit schematic images into accurate circuit netlists using a pipeline that combines object detection with image processing techniques. The first step is component detection and node identification. In this step, we use a YOLO model~\cite{redmon2016you}, trained on circuit symbols, to detect components in the schematic and identify their labels and bounding boxes. To identify the nodes, we use OpenCV’s Connected Components with Stats function~\cite{bradski2000opencv}, which uniquely recognizes all connected wires. To facilitate this, detected components are first removed from the image, leaving only isolated clusters of connected wires that form the nodes. It should be noted that due to the presence of text annotations and other noise in the image, the identified nodes are validated to include only those that intersect the edges of two or more circuit components.

\begin{figure*}[t]
    \centering
    \includegraphics[width=1\textwidth]{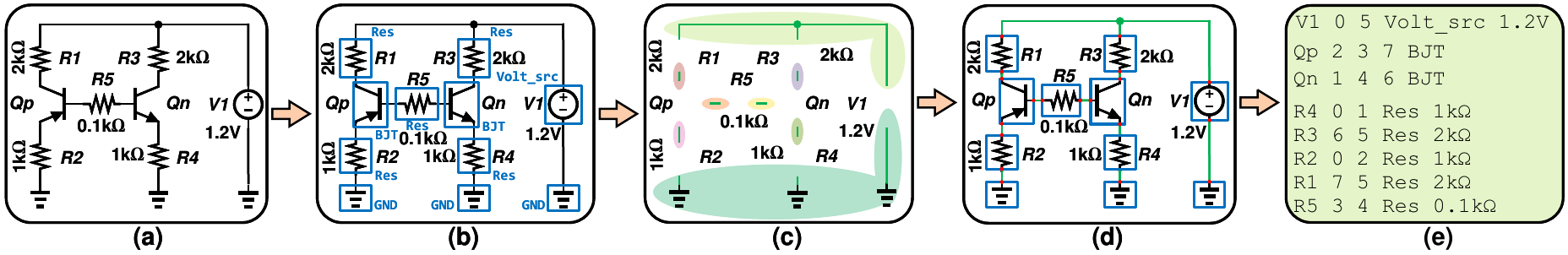}
    \caption{An example of automatic circuit schematic image to netlist generation steps using MuaLLM. (a) The original circuit schematic image. (b) The detection model output with identified components and their bounding boxes. (c) Removing components and clustering nodes. (d) Nodes and their connections to the components. (e) The final generated netlist.}
    \label{fig:netlist_generator}
\end{figure*}

In the next step, for each detected component, we identify which validated nodes are connected to. Each node is assigned a unique numerical identifier, with ground-connected nodes clustered together and assigned the lowest identifier to reflect their shared electrical potential. A visual overview example of this process is shown in Fig.~\ref{fig:netlist_generator}. Using this tool, MuaLLM rapidly creates a large database of circuits, addressing the long-standing challenge of time-consuming labeled data generation through simulations.

\section{Evaluation}
\subsection{Datasets}
Since no publicly available benchmark datasets exist for the evaluation of our task, we curated two custom datasets—RAG-250 and Reas-100—in collaboration with domain experts. These datasets are designed to evaluate the performance of MuaLLM, we curated two datasets: RAG-250 and Reas-100. These datasets are designed to assess the system’s performance in literature-based question-answering and multi-step reasoning tasks related to circuit design. Specifically, we selected research papers on Bandgap Reference (BGR)~\cite{osaki20131} and oscillator circuits~\cite{cao20235} as representative domains to benchmark MuaLLM’s capabilities. While these focus areas provide structured evaluation scenarios, it is important to note that MuaLLM is generalizable and can be applied to any class of circuit literature, enabling broad utility across circuit design workflows.

\subsubsection{RAG-250 Dataset}
This dataset is a collection of question-and-answer (Q\&A) pairs derived from literature on BGR circuits. Since no standard Q\&A dataset exists for the circuit design domain, we manually extracted 250 questions from technical papers. Each question is mapped to expert-verified answers, serving as ground truth labels for evaluating response accuracy. By using this dataset, we assess how well our multimodal RAG retrieves information, synthesizes insights, and generates accurate, literature-backed responses.

\subsubsection{Reas-100 Dataset}
This dataset is designed to evaluate MuaLLM’s multi-step reasoning and analytical capabilities in  circuit design. This dataset, which contains 100 reasoning questions, unlike standard Q\&A datasets~\cite{hotpotqa2025}, focuses on complex  circuit design queries that require deeper reasoning. Answering these queries involves breaking down the problem into logical steps, using relevant tools (\textit{e.g.} \textit{search\_db}, \textit{paper\_fetcher}), and synthesizing a structured response. To evaluate MuaLLM's performance, we manually compare its responses against expert-generated solutions.

By testing both RAG-250 and Reas-100, a comprehensive evaluation of factual accuracy, retrieval efficiency, and reasoning depth of MuaLLM is obtained. Table~\ref{table:sample_both_datasets_queries} shows a sample query from each dataset, demonstrating the types of factual and reasoning-based problems the system is tested on. As it is illustrated in Table~\ref{table:sample_both_datasets_queries}, the RAG-250 example query asks for a direct explanation of a known paper, relying on factual recall. In contrast, the Reas-100 example query requires analyzing design parameters, referencing multiple sources, and synthesizing a solution, demonstrating multi-step reasoning and decision-making. 

\begin {table}[t]
\begin{center}
\begin{threeparttable}
\centering
\caption{Sample queries from RAG-250 and Reas-100 datasets.}
\def\arraystretch{1.2}\tabcolsep 2pt
\label{table:sample_both_datasets_queries}
\begin{tabular}{|>{\centering\arraybackslash}m{12mm}|>{\justifying\arraybackslash}m{74mm}|}
\hline\hline
\multicolumn{1}{|c|}{Dataset} & \multicolumn{1}{|c|}{Example}\\
\hline
\centering \multirow{8}{*}{\centering RAG-250} & \noindent \textbf{Question:} Explain the proposed quadrature voltage-controlled oscillator in the paper "CMOS Quadrature VCOs Using the Varactor Coupling Technique".\\
& \noindent \textbf{Answer:} 
The proposed quadrature voltage-controlled oscillator (QVCO) comprises two cross-coupled LC VCOs coupled via varactors, implemented in TSMC 0.18\,\textmu m CMOS. Each includes an LC tank (inductors and varactors) for tuning and cross-coupled transistors for negative resistance. Varactors enable both tuning and 90\textdegree{} quadrature coupling. Resistors bias the varactors and PMOS bodies. The design operates from 3.86--4.22\,GHz with 3.7\,mW power consumption. The circuit is shown in Fig. 1.\\
\hline
\centering \multirow{13}{*}{\centering Reas-100} & \noindent \textbf{Question:} Design a sub-BGR with a power consumption in the range of 50 nW.\\
& \noindent \textbf{Answer:} To design a sub-BGR with power consumption in the range of 50 nW, I would implement Osaki's sub-BGR architecture as presented in their paper ``1.2-V Supply, 100-nW, 1.09-V Bandgap and 0.7-V Supply, 52.5-nW, 0.55-V Subbandgap Reference Circuits for Nanowatt CMOS LSIs". Their design achieves 52.5 nW power consumption using a nano-ampere current reference circuit, a bipolar transistor, and PTAT voltage generators. The key component is the voltage divider circuit shown in the extracted figure that divides the base-emitter voltage of the bipolar transistor without resistors, enabling sub-1V operation. By following this architecture with the transistor sizing and current reference values specified in their implementation, a sub-BGR with approximately 50 nW power consumption is realized.\\
\hline
\end{tabular}
\end{threeparttable}
\end{center}
\end{table}

\subsection{MuaLLM Results}
\subsubsection{RAG-250 Dataset}
To evaluate the retrieval quality and response accuracy, the system’s answers (using both \texttt{GPT} and \texttt{Claude} as the generative models) are compared against expert-verified ground truth responses. The multimodal RAG achieves an overall recall of \textbf{88.1\%} with \texttt{Claude} and \textbf{90.1\%} with \texttt{GPT}, demonstrating its ability to retrieve and synthesize relevant information effectively. Table~\ref{table:multirag} summarizes the performance evaluation of the multimodal RAG system. The assessment covers overall response accuracy, image citation precision, equation citation recall, and text citation F1 score, showcasing the system's robust performance across multimodal content. The query distribution (the number of each modal queries) is also listed in Table~\ref{table:multirag}.

\begin{table}[t]
\begin{center}
\begin{threeparttable}
\centering
\caption{MuaLLM's multimodal RAG performance evaluation tested with both \texttt{Claude} and \texttt{GPT}.}
\def\arraystretch{1.2}\tabcolsep 2pt
\label{table:multirag}
\begin{tabular}{|M{33mm}|M{15mm}|M{18mm}|M{15mm}|}
\hline\hline
Evaluation Metric & Query Distribution & \multicolumn{2}{|c|}{MuaLLM Multimodal RAG}\\ 
\cline{3-4}
& & \texttt{Claude} & \texttt{GPT}\\
\hline
Overall Response (Recall) & 250 & 88.1\% & 90.1\%\\
\hline
Image Citation (Precision) & 130 & 89.8\% & 93.02\%\\
\hline
Equation Citation (Recall) & 45 & 79.0\% & 73.67\%\\
\hline
Text Citation (F1 Score) & 75 & 93.0\% & 96.0\%\\
\hline
\end{tabular}
\end{threeparttable}
\end{center}
\end{table}

The chosen metrics in Table~\ref{table:multirag}, \textit{i.e.} precision, recall, and F1 score, evaluate key aspects of the multimodal RAG system's performance. Recall assesses completeness for overall responses and equation citations, ensuring relevant information is captured. Precision measures accuracy in image citations, which is crucial for visual data such as circuit diagrams. The F1 score balances precision and recall for text citations, ensuring both accuracy and coverage.

\begin{figure}[t]
    \centering
    \includegraphics[width=1\columnwidth]{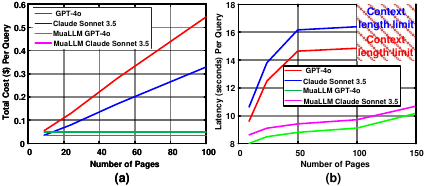}
    \caption{(a) Cost and (b) latency vs pages per query comparison between MuaLLM and the conventional LLMs. The hatched area shows the context length limitation that the conventional LLMs face when processing around more than 100 pages.}
    \label{fig:pages_vs_cost}
\end{figure}

\begin{figure}[t]
\centering
\includegraphics[width=\columnwidth]{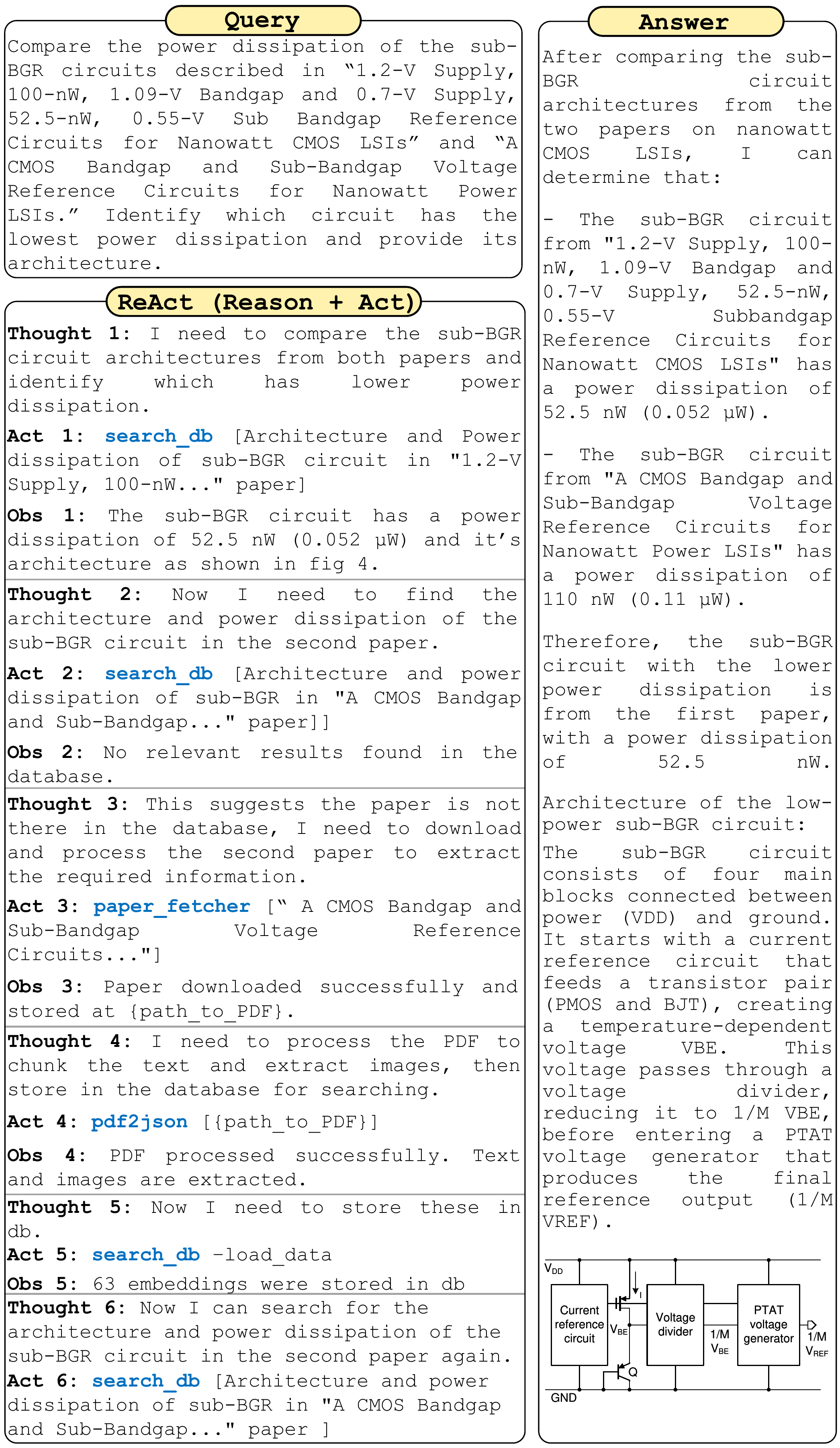}
\caption{Example 1: ReAct (Reason~+~Act) framework for a sample query.}
\label{fig:result_1}
\end{figure}

\begin{figure}[t]
\centering
\includegraphics[width=\columnwidth]{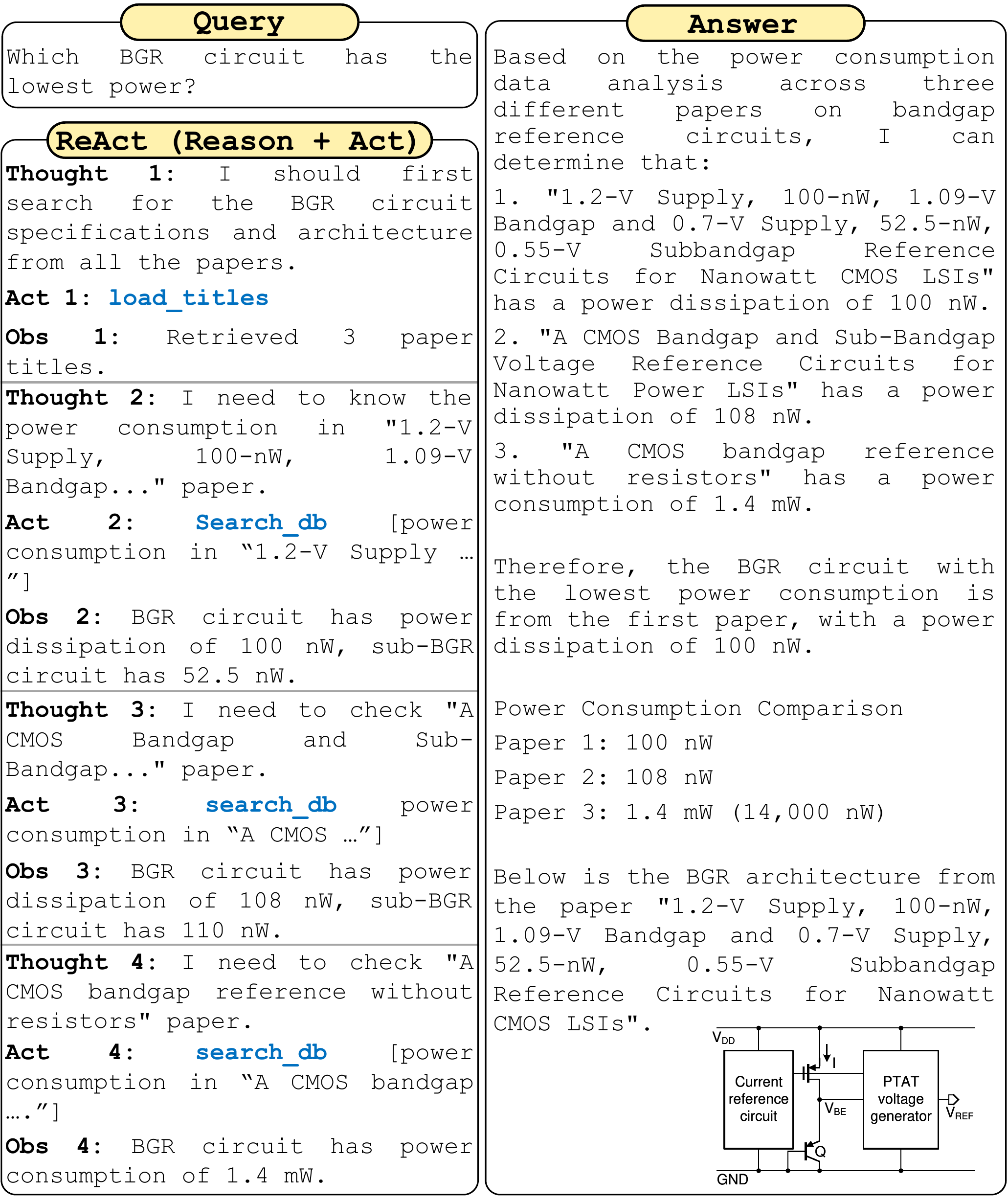}
\caption{Example 2: ReAct (Reason~+~Act) framework for a sample query.}
\label{fig:result_2}
\end{figure}

Since there are no widely available benchmarks to evaluate in terms of scalability and efficiency, we compare MuaLLM against a baseline where full research papers are passed directly as context to standard LLMs (\textit{e.g.} GPT-4o ~\cite{openai2024gpt4o}, Claude Sonnet 3.5 ~\cite{anthropic2024pricing}) for question-answering tasks. While this approach mimics the core objective, retrieving answers from literature, it quickly hits context length limits-fifteen papers~(1,400 tokens/page) saturate most LLM windows ~\cite{anthropic2023promptlength}~\cite{openai2024gpt4o}, making this approach impractical for comprehensive literature reviews. As shown in Fig.~\ref{fig:pages_vs_cost}, cost and latency grow with input page number under the direct full context method. In contrast, MuaLLM retrieves only the most relevant chunks via a hybrid RAG pipeline, decoupling inference from corpus size for cost and computation. At maximum context length, it is at least \textbf{10x} more efficient in cost and \textbf{1.6x} faster while preserving accuracy. One-time preprocessing costs ~\$0.21/paper, enabling low-latency, cost-stable queries suitable for literature-intensive workflows.

While we benchmark our system using direct API-based approaches with \texttt{GPT-4o} and \texttt{Claude Sonnet 3.5}, it is important to note that direct comparisons with existing domain-specific agents such as Ask-EDA~\cite{shi2024askEDA} are not feasible, as it is not open-sourced. Its implementation details are unavailable for evaluation. Meanwhile, Masala-CHAI~\cite{bhandari2024masalaCHAI}, MATLAB Circuit Simulator~\cite{matlab_simulator}, and similar systems focus on synthesis tasks such as SPICE netlist generation from circuit schematics and circuit optimization rather than literature-grounded question answering. Hence, they address different goals and are not directly aligned with the problem MuaLLM solves.

\subsubsection{Reas-100 Dataset}
The reasoning performance of MuaLLM is evaluated by comparing its responses against expert-generated answers. MuaLLM achieves an accuracy of \textbf{84.0\%} with \texttt{Claude} and \textbf{86.8\%} with \texttt{GPT}, demonstrating its ability to provide structured and logical responses. Two examples of the reasoning agent's capabilities are shown in Fig.~\ref{fig:result_1} and Fig.~\ref{fig:result_2}. For demonstration purposes, we illustrate how the workflow would appear if the database contains six papers.

\begin{figure}[t]
\centering
\includegraphics[width=\columnwidth]{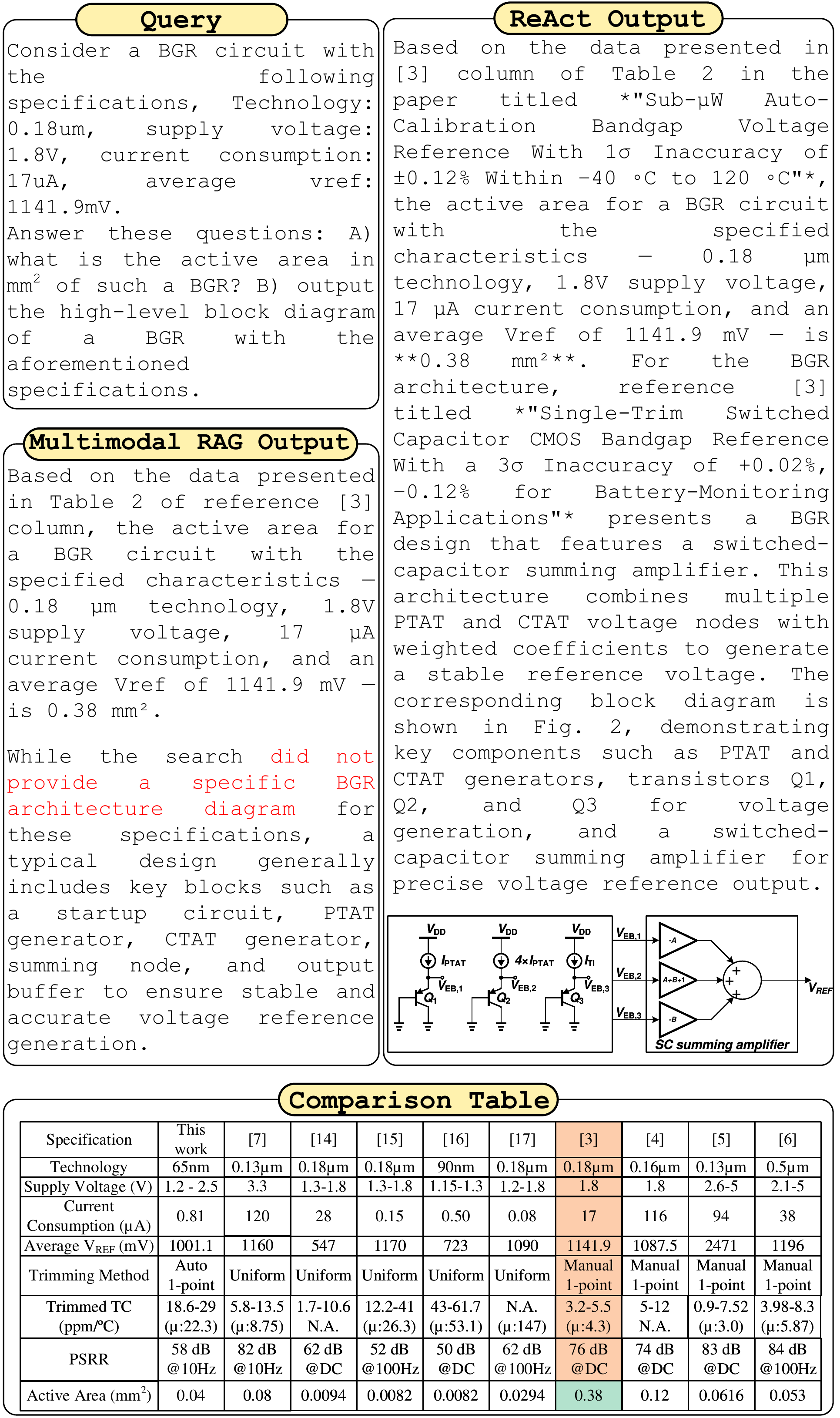}
\caption{A comparison example between outputs of multimodal RAG and ReAct agent to a sample query.}
\label{fig:comparision}
\end{figure}

The ReAct (Reason~+~Act) framework for a sample query is shown in Fig.~\ref{fig:result_1}. This demonstrates an iterative reasoning process where the agent combines logical thinking with strategic actions. It begins by identifying the need to compare two sub-BGR circuits and retrieves available data from the database. Upon discovering incomplete information, the agent reasons that the missing document must be fetched, processed, and stored in the database. Without human intervening,  the agent dynamically adapts by iterating through steps of \textit{reasoning}, \textit{action}, and \textit{observation} until all relevant information is gathered. This adaptive approach enables the agent to efficiently solve complex, multi-step queries with minimal intervention.

In the example shown in Fig.~\ref{fig:result_2}, to compare the power consumption of multiple BGR circuits, the agent first reasons that it must retrieve the titles of all relevant papers in the database, rather than directly searching for power values. This strategic decision ensures the agent has a comprehensive view of all available sources. Once the corresponding paper titles are obtained, the agent iteratively searches for power consumption details in each paper. By methodically examining one source at a time, the agent efficiently handles incomplete information, identifies gaps, and continues searching until all necessary data is gathered. This structured reasoning approach enables the agent to comprehensively analyze the available information and accurately determine which BGR circuit has the lowest power consumption in this example.

Fig.~\ref{fig:comparision} shows a comparison example between the outputs of the Multimodal RAG (MRAG) and ReAct frameworks for the same query. This is to assess their performance in retrieving both numerical values and relevant circuit architectural diagram details. In the case of the MRAG, the system successfully retrieves the required values and the relevant comparison table (Fig.~\ref{fig:comparision} at the bottom~\cite{chi2023sub}) via a direct search. However, since MRAG lacks reasoning capabilities, it fails to recognize that it should further search for the BGR architecture by referring to the paper ``titled [3]" (the corresponding column is highlighted in the comparison table in Fig.~\ref{fig:comparision}). Consequently, MRAG fails to provide the relevant circuit architecture diagram.

In contrast, the ReAct agent demonstrates superior reasoning abilities. After retrieving the comparison table, the agent reasons that it should specifically search the database for the reference ``titled [3]" to obtain the required circuit architecture diagram details. By following this reasoning step, the ReAct agent successfully identifies and retrieves the appropriate BGR architecture diagram. This example highlights the superiority of the ReAct framework in handling complex, multi-step queries. While MRAG retrieves known values, the ReAct agent's ability to plan, adapt, and explore additional references enables it to provide a more complete and accurate response.

\subsection{Netlist Generator Results}
\begin{figure}[t]
\centering
\includegraphics[width=1\columnwidth]{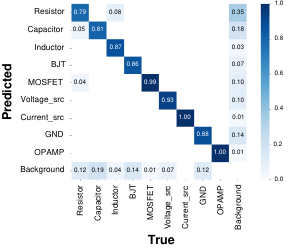}
\caption{Netlist generator component detection confusion matrix.}
\label{fig:confusion_matrix}
\end{figure}

\begin{table}[t]
\begin{center}
\begin{threeparttable}
\centering
\caption{Netlist generator performance evaluation.}
\def\arraystretch{1.2}\tabcolsep 2pt
\label{table:netlist_generator}
\begin{tabular}{|M{12mm}|M{12mm}|M{12mm}|M{45mm}|}
\hline\hline
Precision & Recall & F1 Score & Weighted Mean Average Precision\\
\hline
0.94 & 0.91 & 0.92 & 0.99\\
\hline
\end{tabular}
\end{threeparttable}
\end{center}
\end{table}

The netlist generator is trained and tested on more than 700 and 150 component instances, respectively. The netlist generator performance metrics are summarized in Table~\ref{table:netlist_generator}. Also, the normalized confusion matrix for the component detection is depicted in Fig.~\ref{fig:confusion_matrix}. It should be noted that Masala-CHAI~\cite{bhandari2024masalaCHAI} has not reported any results regarding its schematic image to netlist generation accuracy to compare with.

\section{Conclusion}
In this work, we introduce MuaLLM, an open-source ReAct-based LLM agent with multimodal RAG, to enhance the efficiency of literature review in  circuit design. By integrating iterative reasoning with a hybrid retrieval framework, MuaLLM effectively processes complex multi-objective queries. The agent dynamically adapts to new literature using integrated custom tools such as the Automatic Paper Fetcher, Database Updater, and Netlist Generator. MuaLLM is validated on two custom datasets (RAG-250 and Reas-100). The agent achieves an overall 90.1\% recall on RAG-250. On Reas-100, it reaches 86.8\% accuracy, demonstrating its multimodal citation and multistep reasoning capabilities at scale. At the maximum context length supported by standard LLMs, MuaLLm remains up to 10x less costly and 1.6x faster while maintaining the same accuracy.
\newpage
\bibliographystyle{IEEEtran}

\end{document}